*Article*

# Introducing Various Semantic Models for Amharic: Experimentation and Evaluation with Multiple Tasks and Datasets

Seid Muhie Yimam [1,*], Abinew Ali Ayele [1,2], Gopalakrishnan Venkatesh [3], Ibrahim Gashaw [4] and Chris Biemann [1]

[1] Language Technology Group, Universität Hamburg, Grindelallee 117, 20146 Hamburg, Germany; abinewaliayele@gmail.com (A.A.A.); christian.biemann@uni-hamburg.de (C.B.)
[2] Faculty of Computing, Bahir Dar Institute of Technology, Bahir Dar University, Bahir Dar 6000, Ethiopia
[3] International Institute of Information Technology, Bangalore 560100, India; gopalakrishnan.v@iiitb.org
[4] College of Informatics, University of Gondar, Gondar 6200, Ethiopia; ibrahimug1@gmail.com
* Correspondence: seid.muhie.yimam@uni-hamburg.de; Tel.: +49-4042-883-2418

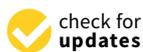

**Abstract:** The availability of different pre-trained semantic models has enabled the quick development of machine learning components for downstream applications. However, even if texts are abundant for low-resource languages, there are very few semantic models publicly available. Most of the publicly available pre-trained models are usually built as a multilingual version of semantic models that will not fit well with the need for low-resource languages. We introduce different semantic models for Amharic, a morphologically complex Ethio-Semitic language. After we investigate the publicly available pre-trained semantic models, we fine-tune two pre-trained models and train seven new different models. The models include Word2Vec embeddings, distributional thesaurus (DT), BERT-like contextual embeddings, and DT embeddings obtained via network embedding algorithms. Moreover, we employ these models for different NLP tasks and study their impact. We find that newly-trained models perform better than pre-trained multilingual models. Furthermore, models based on contextual embeddings from FLAIR and RoBERTa perform better than word2Vec models for the NER and POS tagging tasks. DT-based network embeddings are suitable for the sentiment classification task. We publicly release all the semantic models, machine learning components, and several benchmark datasets such as NER, POS tagging, sentiment classification, as well as Amharic versions of WordSim353 and SimLex999.

**Keywords:** datasets; neural networks; semantic models; Amharic NLP; low-resource language; text tagging

## 1. Introduction

For the development of applications with semantic capabilities, models such as word embeddings and distributional semantic representations play an important role. These models are the building blocks for a number of natural language processing (NLP) applications. Recently, with the advent of more computing power and the widespread availability of a large number of texts, pre-trained models are becoming commonplace. The availability of pre-trained semantic models allows researchers to focus on the actual NLP task rather than investing time in computing such models. In this work, we consider semantic models as the techniques and approaches used to build word representations or embeddings that can be used in different downstream NLP applications.

The work by [1] indicates that word-level representations or word embeddings have played a central role in the development of many NLP tasks. For a named entity recognition task, there are many works that indicate word2Vec lead to a performance boost [2–5]. Static word-embedding models have been also integrated for several NLP tasks such as sentiment analysis [6,7], part-of-speech (POS) tagging [8,9], semantic compositionality [10,11], and many more. While static word-embedding models fail to capture contextual information regarding ambiguous words, the introduction of BERT [12] and similar models have addressed this limitation. The work by [13] indicates that BERT was able to represent the traditional NLP pipeline in an interpretable way, covering some of the basic NLP tasks such as POS tagging, NER, semantic roles, and co-reference resolution.

Even though getting text data is not a problem for low-resource languages, there are only limited efforts in releasing pre-trained semantic models [14,15]. In the case of Amharic, there are very few pre-trained models, for example fastText [16], XLMR [17], and Multi-Flair [18]. Also, these models are produced as part of multilingual and cross-lingual experimental setups, which will not fit the needs of most NLP tasks [14].

In this paper, we have surveyed the existing NLP tasks for Amharic, including available datasets and trained models. Based on the insights on the current state-of-the-art progress on Amharic NLP, we performed different experiments specifically on the integration of semantic models for various tasks, particularly parts-of-speech (POS) tagging, named entity recognition (NER), sentiment analysis, word relatedness, and similarity computation.

The main contributions of this work are many folds: (1) Surveying the existing NLP tasks and semantic models. (2) Computing and fine-tuning nine semantic models and the release of the models publicly along with benchmark datasets for future research. (3) Investigating the main challenges in the computation and integration of semantic models for the Amharic text. (4) Implementing the first Amharic text segmenter and normalizer component and releasing it along with the models and datasets. (5) Release of the word similarity and relatedness datasets (WordSim353 and SimLex999) that have initially been translated using Google Translate API and subsequently have been validated by native speakers. Table 1 shows the different resources (models, tools, datasets) we have contributed. The different strategies and methods used to collect the different dataset and corpus are presented in Section 1.1.2.

**Table 1.** Resources (models, preprocessing tools, and models). For existing resources, we have indicated our contributions. * indicated corpus we have gathered from the web using the Scrapy (https://scrapy.org/ (accessed on 24 October 2021)) open source and collaborative Python framework and using the Tweeter API (https://developer.twitter.com/en/docs/ (accessed on 24 October 2021)). ** indicates the models that we have built using the datasets. *** indicates semantic models that are publicly available.

| Resource | Description | Remark |
| --- | --- | --- |
| NER dataset *** | Benchmark dataset & models | From SAY project |
| POS dataset *** | Benchmark dataset & models | From previous work [19] |
| Sentiment dataset | Different models | Our work [20] |
| word2Vec ** | CBOW and SKipgram | Our corpus * |
| fastText ** | CBOW and SKipgram | Our corpus * |
| fastText *** | CBOW | From fastText |
| DT models ** | Trigram models | Our corpus * |
| XLRM *** | Transformer model | From Huggingface |
| MultFlair *** | Contextualized embedding | From FLAIR repository |
| AmFlair & MultFlairFT ** | Contextualized embedding | Fin-tuned our corpus * |
| AmRoBERTa ** | Transformer model | Newly built our corpus * |
| Pre-processing ** | Tokenization & Segmentation | New tools |
| WordSim & SimLex **** | Word Similarity | Translated from English |

*1.1. Amharic Language*

Amharic is the second most widely-spoken Semitic language (Ethio-Semitic language) after Arabic [21]. It is the working language of the Federal Democratic Republic of Ethiopia and is also the working language of many regional states in the country like Amhara, Ad-dis Ababa, South Nations and Nationalities, Benishangul Gumuz, and Gambella. The language has a considerable number of speakers in all regional states of the country [22].

Amharic is a morphologically-rich language that has its own characterizing phonetic, phonological, and morphological properties [23]. It is a low-resource language without well-developed natural language processing applications and resources.

1.1.1. Pre-Processing and Normalization

Amharic is written in Geez alphabets called Fidel or (ፊደል). In traditional Amharic writing, each word and sentence is supposed to be separated using a unique punctuation mark, namely the Ethiopic comma (፣). However, the modern writing system uses a single space to separate words. Using a single space suffices to split the majority of texts into tokens. However, there are some punctuation marks, such as: (1) The Ethiopic full stop (።) used to mark the end of a sentence. (2) The Ethiopic comma (፣) and the Ethiopic semicolon (፤) that are equivalent to their English counterparts. (3) The Ethiopic question mark (?) that is used to mark the end of questions in Amharic. Moreover, most people also use Latin punctuation marks such as comma, semicolons, question marks, and exclamation marks, even mixing with the Amharic punctuation marks.

For a properly written Amharic text, splitting sentences can be accomplished using the Amharic end of sentence marks (።), question marks, or exclamation marks. However, it might be also the case that people use two Ethiopic commas or two Latin commas to mark the end of a sentence. In the worst case, the Amharic sentence can be delimited with a verb (placed at the end of the sentence) without putting any punctuation marks. As far as we know, there is no proper tool to tokenize words and segment sentences in Amharic. As part of this work, we make available our Amharic segmenter within the FLAIR framework.

Moreover, some of the "Fidels" in Amharic have different representations, for example, the "Fidel" *ሀ* (ha) can have more than four representations (such as ሃ, ሐ, ሓ, ኀ, ኃ, and so on). As the Amharic script originates from the Geez script, the use of different Fidels implied a change in meaning. However, the inherent meaning of the different Fidels became irrelevant in modern writing systems, so users can write with the similar-sounding Fidels interchangeably. These lead to texts written, especially in online communication such as news and social media communications, with different writing styles where the different Fidels are used randomly. For NLP processing, an arbitrary representation of words might pose a serious problem, for example the word ሰው (man) and ሠው (man) might have different embeddings while being the same word. To address this problem, we have built an Amharic text normalization tool that will normalize texts written with different "Fidel" sharing the same sound to a majority class.

1.1.2. Data Sources for Semantic Models

To build distributional semantic models, a large amount of text is required. These days, an enormous amount of texts are being generated continuously from different sources. As we want to build general-purpose semantic models, we collected datasets from different channels, including news portals, social media texts, and general web corpus. For the general web-corpus dataset, we used a focused web crawler to collect Amharic texts. Datasets from the Amharic Web Corpus [24] were also combined to a general-purpose data source. News articles were scraped from January 2020 until May 2020 on a daily basis using the Python Scrappy tool (https://scrapy.org/ (accessed on 24 October 2021)). Similarly, using Twitter and YouTube APIs, we collected tweets and comments written in the 'Fidel' script. In total, 6,151,995 sentences with over 335 million tokens were collected that are used to train the different semantic models.

## 2. Materials and Methods

In this section, we will discuss the pre-trained semantic models and explain the detailed processes we have followed to fine-tune these models. We will also describe the technologies and approaches we have considered to build new models.

*2.1. Distributional Thesaurus*

The distributional hypothesis describes that words with similar meanings tend to appear in similar contexts [25], hence it is possible to build distributional thesaurus (DT) automatically from a large set of free texts. In this approach, if words $w1$ and $w2$ both occur with another word $w3$, then $w1$ and $w2$ are assumed to share some common feature. The more features two words share, the more similar they are considered.

- **AmDT**: The DT was built using the JobImText (http://ltmaggie.informatik.uni-hamburg.de/jobimtext/ (accessed on 24 October 2021)) framework [26]. JoBimText is an open-source framework to compute DTs using lexicalized features that supports an automatic text expansion using contextualized distributional similarity.

*2.2. Static Word Embeddings*

The only pre-trained static word embedding for Amharic text is the **fastText** model, which is trained from Wikipedia and data from the common crawl project [16]. To compare and contrast with the fastText model, we trained a wword2Vec model based on the corpus presented in Section 1.1.2.

- **AmWord2Vec**: Word2Vec [27] helps to learn word representations (word embeddings) that employ a two-layer neural network architecture. Embeddings can be computed using a large set of texts as input to the neural network architecture. The models are built with both the Continuous Bag of Words Model (CBOW) and Skip-gram methods using in 300-dimensional vectors. As seen in Figure 1, the CBOW model considers the conditional probability of generating the central target word from given context words. The Skip-gram approach is the inverse of the CBOW that predicts the context from the target words. We used the Genism Python Library [28] to train the embeddings using the default parameters.
- **fastText**: The pre-trained fastText embeddings distributed by Grave et al. [16] have been trained using a mixture of the Wikipedia and Common Crawl datasets. These 300-dimensional vectors have been trained using Continuous Bag of Words (CBOW) with position-weights, with character n-grams of a length of size 5, a window of size 5, and a negatives sample of size 10.

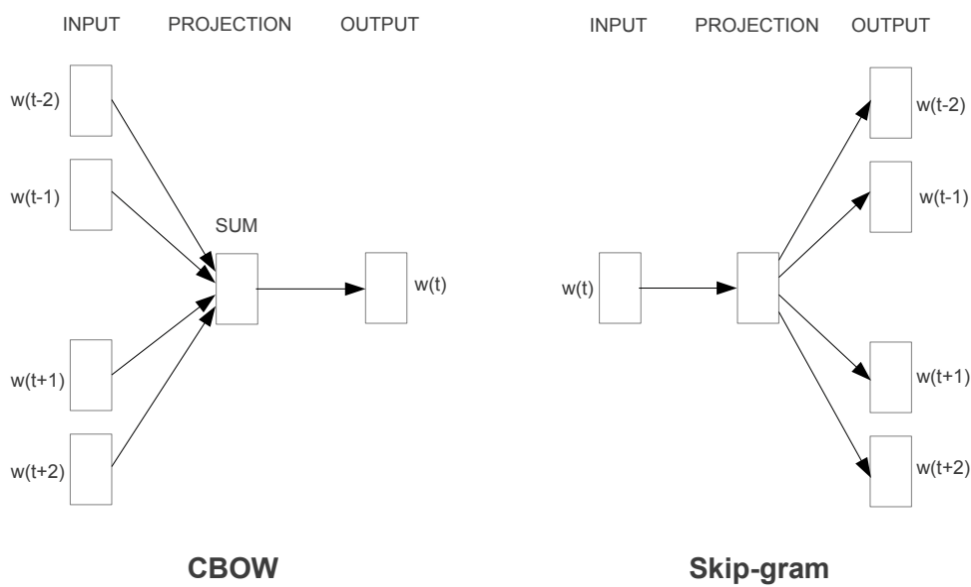

**Figure 1.** Word2Vec model architecture [27].

### 2.3. Network Embeddings

Network embeddings allow representing nodes in a graph in the form of low-dimensional representation (embeddings) to maintain the relationship of nodes [29–31]. In this paper, we first compute the network-based distributional thesaurus (AmDT) and later compute the network embeddings from the DT using DeepWalk [32] and Role2Vec [33] algorithms. These two state-of-the-art network embedding algorithms have been selected for this study as they belong to different categories as explained below.

- **DeepWalk**: The latent node embeddings produced by DeepWalk [32] encodes the social representations, like neighborhood similarity and community membership of graph vertices by modeling a stream of truncated random walks.
- **Role2Vec**: The Role2Vec [33] framework introduces the flexible notion of attributed random walks. This provides a basis to generalize the traditional methods, which rely on random walks, to transfer to new nodes and graphs. This is achieved by learning a mapping function between a vertex attribute vector and a role, represented by vertex connectivity patterns, such that two vertices belong to the same role if they are structurally similar or equivalent. The embeddings have been computed using the karateclub [34] Python library. The network embeddings are trained using the hyperparameter configuration of the package shown in Table 2.

**Table 2.** Training parameters for the different semantic models and NLP applications.

| Model Name | Model Parameters |
| --- | --- |
| DeepWalk | 128 dimensions, walk number 10, walk length 80, window size is 5 |
| Role2Vec | 128 dimensions, walk number 10, walk length 80, window size is 2 |
| MultFlairFT | sequence length of 250, mini batch size of 100, max epochs 10 |
| AmFlair | sequence length of 250, mini batch size of 100, max epochs 10 |
| AmRoBERTa | epochs of 5, per gpu train batch size of 8, block size of 512 |
| SequenceTaggers | hidden size of 256, mini-batch size of 32, epochs of 150 |

### 2.4. FLAIR Embeddings

FLAIR embeddings are contextualized embeddings, which are trained based on sequences of characters where words are contextualized by their surrounding texts [35]. Unlike word2Vec embeddings, FLAIR embeddings enable us to compute different representations for the same word based on the surrounding contexts, as shown in Figure 2. In addition to the contextualized word-embeddings computation, the FLAIR framework integrates document embedding functionalities such as *DocumentPoolEmbeddings*, which produces document embeddings from pooled word embeddings and *DocumentLSTMEmbeddings*, which provides document embeddings from LSTM over word embedding [36]. For this experiment, we have considered three semantic models based on the FLAIR contextual string embeddings.

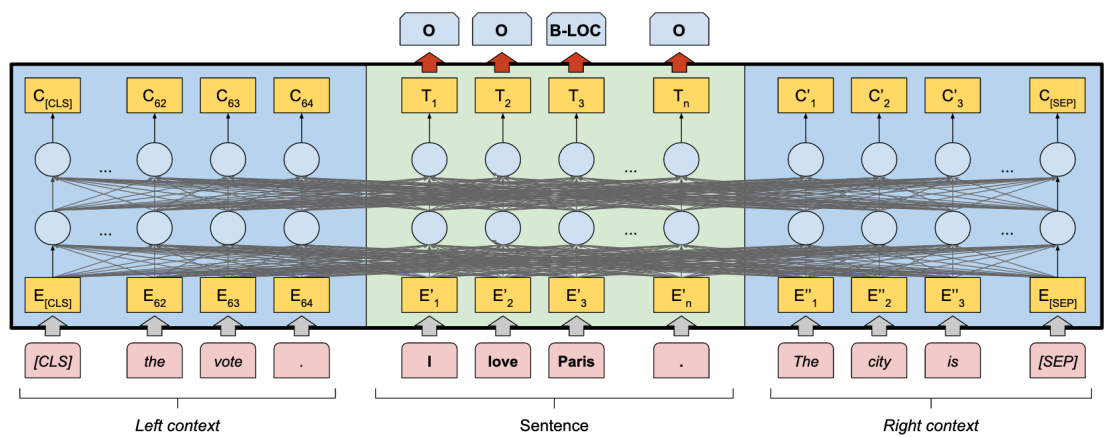

**Figure 2.** The FLAIR model architecture [37] for document-level features.

- **MultFlair** : As part of the FLAIR embedding models ecosystem, Schweter [18] has built multilingual word embedding using the JW300 corpus [38]. JW300 is compiled from parallel corpora of over 300 languages with around 100,000 parallel sentences per language pair.
- **MultFlairFT**: We have fine-tuned the MultFlair embedding model using our corpus. Fine-tuning the model runs on our GPU server, which was completed in 18 days.
- **AmFlair**: This is a new FLAIR embedding model we have trained from scratch using our corpus. The training is performed on a GPU server (GeForce RTX 2080) with the training parameters shown in Table 2. The training was completed in 6 days.

### 2.5. Transformer-Based Embeddings

With the release of Google's Bidirectional Encoder Representations from Transformer (BERT) [12], word representation strategies have shifted from the traditional static embeddings to a contextualized embedding representation. While Figure 3 shows the transformer model architecture [39], Figure 4 shows the pre-training and fine-tuning procedures in BERT. BERT-like models have an advantage over static embeddings as they can accommodate different embedding representations for the same word based on its context. In this task, we have used RoBERTa, which is a replication of BERT developed by Facebook [40]. Unlike BERT, RoBERTa removed the *next sentence prediction* functionality to train on longer sequences, dynamically changing the masking patterns.

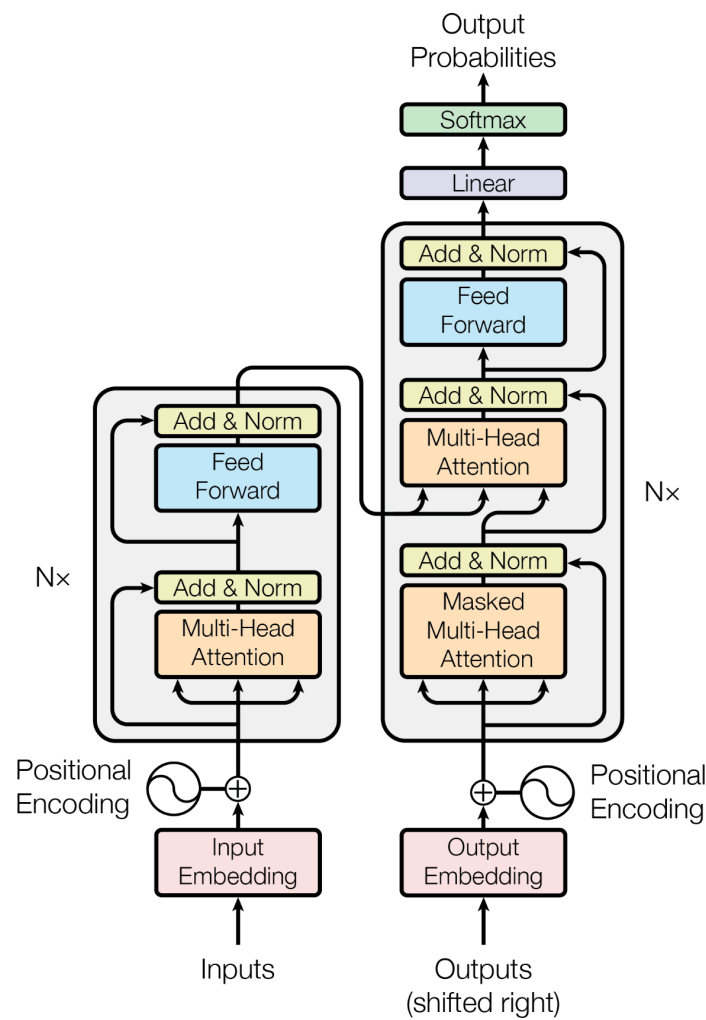

**Figure 3.** The transformer model architecture [39].

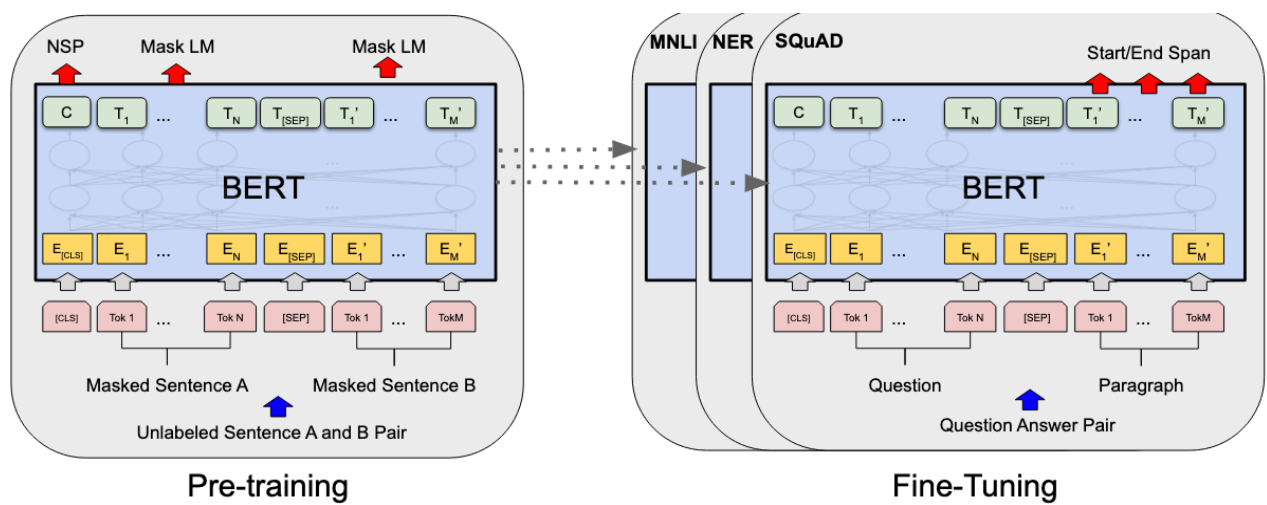

**Figure 4.** Architecture of the BERT pre-training and fine-tuning procedures [12].

In this experiment, two transformer-based embedding models are used.

- **XLMR**: Unsupervised Cross-lingual Representation Learning at Scale (XLMR) is a generic cross-lingual sentence encoder that is trained on 2.5 TB of newly-created clean CommonCrawl data in 100 languages including Amharic [17].
- **AmRoBERTa**: Is a RoBERTa model that is trained using our corpus, as discussed in Section 1.1.2. It has been trained using 4 GPUs (Quadro RTX 6000 with 24GB RAM) and it has taken 6 days to complete, with parameters shown in Table 2.

## 3. Results

In this section, we will report the results for different NLP tasks using the existing and newly-built semantic models. We have also compared the differences in using manually-crafted features and embeddings for machine-learning components.

### 3.1. Most Similar Words and Masked Word Prediction

One of the most prominent operations to perform using static Word2Vec embeddings is to determine the most similar *n* words for a target word.

As seen from Table 3, most of the top *n* similar words from fastText are of a bad quality. We observed that this is due to the fact that the text extracted from Wikipedia is smaller in size so that the word occurs in very few sentences. For the word "ox", the top prediction is a wrong candidate that is instantly retrieved from the first entry in Wikipedia, which is a figurative speech (https://bit.ly/2Beuzi2 (accessed on 24 October 2021)).

**Table 3.** Comparison of word similarities computed using the pre-trained fastText and the AmWord2Vec models. The English glossaries are an approximate as some of the translations will be very long to put in the table, for example, the word **አይከናወኑለትም** can be translated as "they can not be performed for him".

| በሬ(ox) | | መብላት(eating) | |
|---|---|---|---|
| **fastText** | **AmWord2Vec** | **fastText** | **AmWord2Vec** |
| ባላገደደ(tilted) | ፍየል (goat) | ካማራት (want) | መመገብ (feed) |
| አየሀና (see) | ወይፈን (bull) | አይደፍርም (untouchable) | መግዛት (buy) |
| ይፈንዳ (explodee | ዶሮ (hen) | አይከናወኑለትም (not perform) | ማጠጣት (drink) |
| ያበጠው (swollen) | በለቅ (donkey) | መጠጣትና (drink) | ማሽተት (smell) |
| ባልገባ (enter) | ሙክት (goat) | ዳቦን (bread) | መደነስ (dance) |
| ለሚጎተት (drag) | በግና (sheep) | ፍሬና (fruit) | መሸጥ (sell) |
| ቀንዱን (horn) | በግ (sheep) | ላምና (cow) | መሽናት (urinate) |

**Table 3.** *Cont.*

| በሬ(ox) | | መብላት(eating) | |
|---|---|---|---|
| **fastText** | **AmWord2Vec** | **fastText** | **AmWord2Vec** |
| ካራጁ (killer) | አህያ (donkey) | የምችል (can) | መጋገር (backe) |
| አትከልክለው (prohibit) | እደተናገረው (said) | በማጥቃትና (attack) | መቁጠር (count) |
| ቀንዳምን (horny) | ጥጃ (calf) | መብላትና (eat) | መሸመት (buy) |
| ተሴት (female) | ቆዳውን (leather) | በእንጀራ (Injera) | መጋት (drink) |
| አንኮሌ (Fool) | ሰንጋ (ox) | መጥገብ (satisfied) | መጨፈር (dance) |
| ተላም (cow) | ሲያርስ (plow) | እንደማይወድ (no like) | ለመብላት (eat) |
| ገደሉን (downhill) | ለምዱን (sheepskin) | የብይ (marbles) | ማጠብ (wash) |
| ለቀንዳም (horny) | ጅብ (hyena) | ያቃተን (unable) | ማሰብ (think) |

The BERT-like transformer-based embeddings such as RoBERTa and XLRM also support predicting the *n* most probable words to fill by masking an arbitrary location in a sentence. As shown in Table 4, we compare the results suggested by AmRoBERTa and the suggestions provided by AmDT and AmWord2Vec models. To contrast the predictions using AmRoBERTa, we present the two sentences that are shown in Examples 1 and 2, where we mask a context-dependent word ትርፍ, which can be considered as "profit" in the first sentence and "additional" in the second sentence.

**Table 4.** Comparison of similar words generated from the AmDT, AmWord2Vec, and two contextualized suggestions from AmRoBERTa for the word ትርፍ. Columns AmRoBERTaS1 and AmRoBERTaS2 show the the contextual suggestion for the <mask> word from Sentence1 and Sentence2 of Examples 1 and 2.

| ትርፍ: 1. profit, 2. additional | | | |
|---|---|---|---|
| **amDT** | **amWord2Vec** | **AmRoBERTaS1** | **AmRoBERTaS2** |
| ገቢ (income) | ገንዘብ (money) | ገንዘብ (money) | አንድ (one) |
| ጥቅም (advantage) | ገቢ (income) | ገቢ (income) | የሆነ (is) |
| እርካታ (statsfaction) | ጥቅም (advantage) | እድል (chance) | ብዙ (many) |
| ውጤት (result) | ዋጋ (price) | ዋጋ (price) | ማንኛውም (any) |
| ተቀባይነት (acceptance) | ፍጆታ (consumption) | ብር (money) | ሁሉም (all) |
| ምንዛሪ (exchange) | ትርፉን (additional) | ድጋፍ (support) | ሌላ (other) |
| ስኬት (success) | ጠቀሜታ (advantage) | ስራ (work) | ሁለት (two) |
| መፍትሄ (solution) | ምርት (product) | ሀብት (wealth) | ትልቅ (large) |
| ፋይዳ (advantage) | ገቢም (income) | አቅም (power) | ማንም (anyone) |
| እፎይታ (relief) | ዋጋም (price) | ድርሻ (share) | ሶስት (three) |
| ደስታ (happiness) | ፋይዳ (advantage) | ጥቅም (advantage) | አንድም (one) |
| ፈውስ (medicament) | ገቢው (income) | ግብር (tax) | አብዛኛው (many) |

> **Example 1.** : በተለይም ነጋዴዎች በትንሽ ወጪ ብዙ <mask> የማግኘት ዓላማ በመያዝ ውኃን ከወይን ጋር መደባለቅ... Particularly Merchants, to get more <mask> with less expenditure by mixing water with Wine ...
>
> **Example 2.** : ታክሲዎችና ባጃጆች ከተፈቀደው መጠን በላይ <mask> ሰው መጫን ካላቆሙ ከስራ ውጪ ይደረጋሉ ተባለ። If Taxis and Bajajs do not stop transporting <mask> people than allowed they will be out of a job.

### 3.2. Word Similarity and Relatedness Tasks

WordSim353 (http://alfonseca.org/eng/research/wordsim353.html (accessed on 24 October 2021)) and SimLex999 (https://fh295.github.io/simlex.html (accessed on 24 October 2021)) are datasets developed to measure semantic similarity and relatedness between terms [41]. WordSim measures semantic relatedness on a rating scale while SimLex is specifically designed to capture the similarity between terms [42]. Word similarity and relatedness can be measured using word embeddings and context embeddings [43,44]. As we do not have these resources for Amharic, we have used the English WordSim353 and SimLex999 datasets to construct the similarity and relatedness resources. To construct the datasets, we translate the WordSim353 and SimLex999 dataset from English to Amharic using the Google translate API. Since the Google translate API for Amharic is not accurate enough, the dataset is verified by two native Amharic speakers. We removed wrongly translated word pairs and multiword expressions from the dataset. These datasets are one of the contributions of this work that will be published publicly.

We have used the different semantic models to measure the similarity and relatedness scores based on the existing benchmark approaches. The experimental setup follows the established strategy of computing the Spearman correlation ($\rho$) between the cosine similarity of the word vectors or embeddings and the ground truth score [43]. Table 5 presents the results from this quantitative evaluation.

**Table 5.** Spearman correlation ($\rho$) and standard deviation ($\sigma$) scores on the Amharic Wordsim353 and SimLex999 datasets.

|  | **Spearman Correlation ($\rho$)** |  | **std ($\sigma$)** |  |
| --- | --- | --- | --- | --- |
| **Models** | **Wordsim353** | **SimLex999** | **Wordsim353** | **SimLex999** |
| AmWord2Vec | 0.518 | 0.285 | 0.247 | 0.274 |
| fastText | 0.434 | **0.314** | 0.238 | 0.245 |
| AmFlair | 0.444 | 0.288 | 0.183 | 0.208 |
| MultFlairFT | 0.447 | 0.272 | 0.166 | 0.189 |
| MultFlair | 0.173 | 0.231 | 0.085 | 0.109 |
| AMRoBERTa | 0.285 | 0.202 | 0.141 | 0.133 |
| XLMR | 0.182 | 0.183 | 0.075 | 0.065 |
| DeepWalk | **0.523** | 0.191 | **0.279** | **0.308** |
| Role2Vec | 0.448 | 0.255 | 0.202 | 0.233 |
| English datasets state-of-the-art |  |  |  |  |
|  | 0.828 [45] | 0.76 [46] | - | - |

From Table 5, we can see that the "DeepWalk" model works better for the WordSim353 dataset while "AmFlair" and "AmWord2Vec" works better for the SimLex999 datasets. Furthermore, the newly-trained as well as the fine-tuned models produce a better result than the pre-trained embeddings ("XLMR" and "MultFliar"). The low standard deviation ($\sigma$) results are due to the fact that most of the similarity scores (cosine similarity) between the "word1" and "word2" embeddings are higher. The higher similarity score between the two words is caused by having almost identical embeddings for each of the words, as the embeddings might not be optimized towards the specific tasks. However, we suggest further investigation to check the quality of the embeddings in the pre-trained models. Please note that the results are not directly comparable with the English datasets (Table 5 at the bottom) as we have kept entries where we have direct translation and when the translation does not lead to multi-word expressions.

Moreover, we have conducted some error analysis on the similarity and relatedness results. The following are some of the observations identified.

1. **Translation errors**: We found out that some of the translations are not accurate. For example, the word pairs 'door' and 'doorway' received the same translation in Amharic as "በር".
2. **Equivalence in translation**: The other case we have observed is that some similar/related words received the same translations. For example, the word pairs 'fast' and 'rapid', both are translated as "ፈጣን".

3. **SimLex antonym annotation**: The ground truth annotation scores for the word pairs 'new' and 'ancient' as well as 'tiny' and 'huge' are near zero. However, the cosine similarity produces a negative result as the word pairs are opposite in meaning.
4. **Pre-trained vs. fine-tuned models**: Finetuning the pre-trained models results in a higher similarity (Table 5). For the word pairs 'professor' and 'cucumber' in WordSim353, the ground truth score is 0.31 (10 is a maximum score). The pre-trained 'MultFlair' model results in a wrong (which is higher) similarity score (0.819) but the fine-tuned model results in a smaller score (0.406) near to the ground truth score.

We can also observe that the scores for SimLex999 are lower than the WordSim353 scores. The work by Hill et al. [44] indicated that LexSim999 tasks are challenging for computational models to replicate as the model should capture similarity independently of relatedness association. Moreover, the results on both datasets for Amharic are lowered compared with the English datasets. This can be attributed to several reasons such as: (1) As the translations are not perfect, the ground truth annotation scores should not be used as it is for Amharic pairs, and (2) as Amharic is morphologically complex, the embeddings obtained might be different from the correct lemma of the word. Hence, the pairs should be first lemmatized to the correct dictionary entry before computing the similarity scores. Furthermore, the annotation should be done by Amharic experts to calculate the similarity and relatedness scores.

### 3.3. Parts-of-Speech Tagging

Using a corpus of one-page long Amharic text, Getachew [47] has developed a part of speech tagger using the stochastic Hidden Markov Model approach that can extract major word classes noun, verb, adjective, auxiliary), but fails to determine subcategory constraints such as number, gender, polarity, tense case, definiteness, and unable to learn from new instances. Gambäck et al. [48] have developed an Amharic POS tagger using three different tag-sets. Similarly, Tachbelie and Menzel [49] have prepared 210,000 manually annotated tagged tokens and developed a factored language model using SVM and HMM. Tachbelie et al. [50] have employed a memory-based tagger, which is appropriate for low-resourced languages like Amharic.

Despite the several works on Amharic POS tagging, there is, as far as we know, no publicly available POS tagger model and benchmark dataset that can be used for downstream applications.

For our Amharic POS tagging experimentation, we have trained different POS tagging models using the dataset compiled by Gashaw and Shashirekha [19]. The dataset is comprehensive in the sense that texts from the different genres are incorporated. The dataset from the Ethiopian Language Research Center (ELRC) of Addis Ababa University [51] consists of news articles covering different topics such as sport, economics, politics, etc. amounting to a total of 210,000 words. The ELRC dataset supports 11 basic tags. The work by Gashaw and Shashirekha [19] extends the ELRC dataset by annotating texts from Quranic and Biblical texts (called ELRCQB). In total, 39,000 sentences are annotated for Amharic POS tags using 62 tags. The dataset is split into training, development and testing instances. The development set is used to optimize parameters during training.

Table 6 shows the experimental results using different models. We built two types of models, the Conditional Random Field (CRF) tagging model and the sequence classifier model using the FLAIR deep learning framework. For the CRF model, the "sci-kit-learn-crfsuite" python wrapper of the original CRF implementation is employed (Model **CRF-Suite** in Table 6). The features for the CRF model include (1) the word, previous word, and the next word, (2) prefixes and suffixes with lengths ranging from 1 to 3, and (3) checking if the word is numeric or not. The remaining models are built using the FLAIR sequence classifier based on the different semantic models using the parameters shown in Table 2 with a GPU:GeForce GTX 1080 11GB server.

**Table 6.** Experimental results for POS tagging (macro-averages). * indicated the stat-of-the-art result of the ELRCQB dataset but the results are not comparable as: (1) The reported result shows only the accuracy, and (2) they are not benchmark datasets.

| **Model** | **Precision** | **Recall** | **F1** |
|---|---|---|---|
| CRFSuite Word Features | | | |
| CRFSuite | **94.78** | **94.81** | **94.74** |
| Word Embeddings | | | |
| AmWord2Vec | 81.41 | 81.94 | 81.05 |
| fastText | 84.18 | 84.46 | 83.94 |
| Contextual Embeddings | | | |
| AmRoBERTa | 94.08 | 94.13 | 94.08 |
| AmFlair | 91.75 | 91.71 | 91.69 |
| MultFlairFT | 91.19 | 91.07 | 91.06 |
| XLMR | 94.20 | 94.17 | 94.16 |
| MultFlair | 89.65 | 89.53 | 89.48 |
| Graph Embeddings | | | |
| DeepWalk | 82.27 | 82.71 | 81.81 |
| Role2Vec | 81.45 | 81.95 | 80.91 |
| State-of-the-art result | | | |
| 92.27 accuracy * | | | |

### 3.4. Named Entity Recognition

Named entity recognition (NER) is a process of locating and categorizing proper nouns in text documents into predefined classes like a person, organization, location, time, and numeral expressions [52]. In this regard, one of the early attempts for Amharic NER is the work by Ahmed [53], which is conducted on a corpus of 10,405 tokens. Using the CRF classifier, the research indicated that POS tags, suffixes, and prefixes are important features to detect Amharic named entities. The work by Alemu [54] also conducted similar work on a manually developed corpus of 13,538 words with the Stanford tagging scheme. The work by Tadele [55], an approach for a hybrid Amharic named entity recognition employed a combination of machine learning (decision trees and support vector machines) and rule-based methods. They have reported that the pure machine learning approaches with POS and nominal flag features outperformed the hybrid approach.

Another work by Gambäck and Sikdar [56] also developed language-independent features to extract Amharic named entities using a bi-directional LSTM deep learning neural network model and merged the different feature vectors with word embedding for better performance. Sikdar and Gambäck [57] later employed a stack-based deep learning approach incorporating various semantic information sources that are built using an unsupervised learning algorithm with word2Vec, and a CRF classifier trained with language-independent features. They have reported that the stack-based approach outperformed other deep learning algorithms.

The main challenges with Amharic NER are: (1) As there are no capitalization rules in the language, it is very difficult to build a simple rule or pattern to extract named entities. (2) POS tags might help in discriminating named entities from other tokens, however, there is no publicly-available POS tagger for Amharic that can be integrated into NER systems. (3) Most of the research is carried out as part of academic requirements for a Bachelor's and Master's thesis where the research output was not well documented. Moreover, there are no benchmark dataset or tools that could advance future research in Amharic. In this work, we explore the effectiveness of semantic models for Amharic NER and release both the NER models and the benchmark datasets publicly (in Table 7).

**Table 7.** Experimental results for NER classification (macro-averages).

| Model | Precision | Recall | F1 |
|---|---|---|---|
| *CRFSuite Word Features* | | | |
| CRFSuite | 80.88 | 63.89 | 71.39 |
| *Word Embeddings* | | | |
| AmWord2Vec | 68.11 | 57.47 | 62.34 |
| fastText | 79.01 | 55.56 | 65.24 |
| *Contextual Embeddings* | | | |
| AmRoBERTa | 25.62 | 30.56 | 27.87 |
| AmFlair | 79.70 | 74.31 | 76.91 |
| MultFlairFT | **81.21** | **74.31** | **77.61** |
| MultFlair | 75.58 | 61.81 | 68.00 |
| *Graph Embeddings* | | | |
| DeepWalk | 73.95 | 64.06 | 68.65 |
| Role2Vec | 74.00 | 60.76 | 66.73 |

For this experiment, the Amharic dataset annotated within the SAY project at New Mexico State University's Computing Research Laboratory was used. The data is annotated with six classes, namely person, location, organization, time, title, and others. There are a total of 4237 sentences where 5480 tokens out of 109,676 tokens are annotated as named entities. The dataset is represented in XML format (for the different named entity classes) and is openly available in GitHub (https://github.com/geezorg/data/tree/master/amharic/tagged/nmsu-say (accessed on 24 October 2021)). The same approach as the POS tagger systems was used to train the NER models.

*3.5. Sentiment Analysis*

The task of sentiment analysis for low-resource languages like Amharic remains challenging due to the lack of publicly available datasets and the unavailability of required NLP tools. Moreover, there are no attempts of analyzing the complexities of sentiment analysis on social media texts (e.g., Twitter dataset), as the contents are highly context-dependent and influenced by the user experience [58]. Some of the existing works in Amharic either target the generation of sentiment lexicon or are limited to the manual analysis of very small social media texts.

The work of Alemneh et al. [59] focuses on the generation of Amharic sentiment lexicon using the English sentiment lexicon. The English lexicon entries are translated to Amharic using a bilingual English-Amharic dictionary.

The work by Gebremeskel [60] builds a rule-based sentiment polarity classification system. Using movie reviews, 955 sentiment lexicon entries are generated. The system is built to detect the presence and absence of the positive and negative sentiment lexicon entries to classify the polarity of the document.

For this work, we considered the recently released sentiment classification datasets, a total of 9400 k tweets where each tweet is annotated by three users [20]. The tweets are sampled using the extended sentiment lexicons from Gebremeskel [60] (to a total of 1194 lexicon entries). We split the dataset into training, testing, and development sets with the 80:10:10 splitting strategy.

We built document-based sentiment classifiers using the different semantic models. For a classical classification approach, we used the term frequency-inverse document frequency (TF-IDF) features using different algorithms from the sci-kit-learn Python machine learning framework. For the deep learning approach, we used the TextClassifier document classification model from the FLAIR framework, which uses the *DocumentRNNEmbeddings* computed from the different word embeddings. Table 8 shows the different experimental results.

**Table 8.** Experimental results of the test sets on the sentiment classes: "Positive", "Negative" and "Neutral" (macro-averages).

| Model | Precision | Recall | F1 |
|---|---|---|---|
| TF-IDF representation | | | |
| LogReg | 46.80 | 60.88 | 52.92 |
| RanfomF | 44.59 | 52.17 | 48.09 |
| KNN | 49.85 | 50.22 | 50.03 |
| NearestC | 47.36 | 49.20 | 48.26 |
| SVM | 35.44 | 46.42 | 40.20 |
| Word Embeddings | | | |
| AmWord2Vec | 55.54 | 54.91 | 55.22 |
| fastText | 43.50 | 67.81 | 53.00 |
| Contextual Embeddings | | | |
| AmFlair | 53.24 | 59.25 | 56.09 |
| MultFlair | 46.96 | 55.05 | 50.68 |
| MultFlairFT | 54.49 | 59.58 | 56.92 |
| AmRoBERTa | 46.62 | 56.39 | 51.04 |
| Graph Embeddings | | | |
| DeepWalk | 55.89 | 57.71 | 56.78 |
| Role2Vec | **56.26** | **60.89** | **58.48** |
| Baselines | | | |
| Stratified | 33.79 | 33.80 | 33.80 |
| Uniform | 29.72 | 30.96 | 30.33 |
| MostFreq | 33.33 | 17.31 | 22.78 |

## 4. Discussion

In this section, we will briefly discuss the effects of the different semantic models for the respective NLP tasks.

- **Top *n* similar words**: For the top *n* similar words experiment, as seen from Table 3, the fastText model seems to produce irrelevant suggestions compared to the AmWord2Vec model, which is associated with the smaller corpus size used to train fastText. If we search the predicted word provided by fastText on the Amharic Wikipedia page, we can see that the word co-occurs with the target word only on one occasion. However, as we did not train a new fastText model from scratch using our dataset, we can not claim if the suggestions are mainly due to the size of the dataset or due to the training approach used in fastText architecture.

  The similar words suggested by AmDT, AmWord2Vec, and DeepWalk are comparable (see Table 3 and 9). In general, the suggestions from AmDT are more fitting than the ones generated using AmWord2Vec, but the suggestions from AmDT and DeepWalk are equally prevalent to the target words. This corresponds to the finding that the conversion of DTs to a low-level representation can easily be integrated into different applications that rely on word embeddings [61].

**Table 9.** Comparison of similar words generated from the AmDT and network embedding (Deepwalk) representations.

| በሬ(ox) | | መብላት(eating) | |
| --- | --- | --- | --- |
| **AmDT** | **DeepWalk** | **AmDT** | **DeepWalk** |
| በሬ (ox) | ፍየል (goat) | መብላት (eating) | መመገብ (feeding) |
| በግ (sheep) | በግ (sheep) | መመገብ (feeding) | መጠጣት (drinking) |
| ከብት (cattle) | ከብት (cattle) | መኖር (living) | መተኛት (sleeping) |
| ዶሮ (hen) | እንስሳ (animal) | መጫወት (playing) | መሸጥ (selling) |
| ፍየል (goat) | ጅብ (hyena) | መስራት (working) | ማደር (sleeping) |
| እንስሳ (animal) | አይጥ (mouse) | መስጠት (giving) | መግዛት (buying) |
| ፈረስ (horse) | እባብ (snake) | መሸጥ (selling) | ማምረት (producing) |
| አህያ (donkey) | ዝሆን (elephant) | መነጋገር (talking) | ማስቀመጥ (putting) |
| ሰንጋ (steer)( | ሬሳ (corse) | መንቀሳቀስ (moving) | መውሰድ (taking) |
| በሬዎች (oxen) | ዶሮ (hen) | መግባት (entering) | ማልቀስ (crying) |
| ሰው (man) | እንቁላል (egg) | ማከናወን (accomplishing) | መትከል (planting) |
| እባብ (snake) | ስጋ (meat) | መሰብሰብ (collecting) | መተንፈስ (breazing) |
| ስጋ (meat) | ቅቤ (butter) | መጻፍ (writing) | መጥራት (calling) |
| ላም (cow) | ፈረስ (horse) | መጠጣት (drinking) | መጫወት (playing) |
| እርሻ (farm) | ውሻ (dog) | መንዝ (traveling) | መቁጠር (counting) |
| አንበሳ (lion) | አህያ (donkey) | መጠቀም (using) | መጮህ (screaming) |
| ውሻ (dog) | ስጋውን (meat) | መውጣት (going out) | ማውጡቱ (taking out) |
| በሬውን (ox) | ሳር (grass) | መውሰድ (taking) | መተው (leaving) |
| ካህን (pastor) | ዝንጀሮ (monkey) | ማገልገል (serving) | ማተኮር (concentrating) |
| በበሬ (ox) | ዛፍ (tree) | መምራት (leading) | መገናኘት (meeting) |

However, the similar words predicted by the DT and word2Vec embeddings are static and it is up to the downstream application to discern the correct word that fits the context. However the next word prediction using the transformer-based models, in this case from the AmRoBERTa model, predicts words that can fit the context. From Examples 1 and 2, we can see that the "masked" words are to be predicted in the two sentences. The two sentences are extracted from the online Amharic news channel (https://www.ethiopianreporter.com/ (accessed on 24 October 2021)) and we masked the word ትርፍ in both sentences. In the first sentence (S1), it refers to "profit" while in the second sentence (S2), it intends "additional" or "more". We can see from Table 4 that the AmRoBERTa model generated words that can fit the context of the sentence. We have also observed that AmRoBERTa helps in word completions tasks, which is particularly important for languages such as Amharic, as it is morphologically complex.

- **Word similarity/relatedness**: For the WordSim353 word pair similarity/relatedness experiment, the "DeepWalk" and "AmWord2Vec" models produce the best results. While this is the first dataset and experiment for the Amharic, the Spearman's correlation ($\rho$) result is better than most of the knowledge-based results for the English counterpart datasets. The state-of-the-art result for English reaches a score of 0.828, which is much larger than the results for Amharic scores. We could not compare the results for several reasons such as (1) errors that occurred during translation and (2) the ground truth annotation scores are directly taken from the English dataset, which might not be optimal. In the future, we suggest to re-annotate the datasets using Amharic experts. We have also observed that the Simlex999 datasets are challenging for the similarity computation task. The "fastText" model achieves better results compared to the other Amharic semantic models. Pre-trained models achieve the lowest results as we can witness from the lower standard deviation ($\sigma$) scores.
- **POS tagging**: As seen in Table 6, the semantic models from the transformer-based embeddings perform as good as the CRF algorithm for the POS tagging task. While training the deep learning models took much longer than the CRF algorithm, using deep learning models avoids the need to identify features manually. Both AmRoBERTa and CRF-based models predict the conjunctions, interjections, and prepositions correctly. However, for the rare tags (that occurs fewer times), such as ADJPCS (Adjective with prep. & conj. singular) and VPSs (Verb with prep. singular), AmRoBERTa predicts the

- nearest popular tag. However it was observed that CRF perfectly memorizes the correct tag. For example, the word **ኢንድያገኑ**, which should be a spelling error (maybe the last *s* stands for spelling error in VPSs ), is tagged as VPS with AmRoBERTa. We have also observed that the FLAIR contextual embeddings perform very well compared to the network embedding models. The new Amharic FLAIR embeddings (**AmFlair**) and the fine-tuned models (**MultFlairFT**) produce a slightly better result than the publicly-available multilingual FLAIR (**MultFlair** embeddings).
- **Named entity recognition**: In the case of the named entity recognition task, the transformer model performs poorly compared to the CRF and FLAIR embedding models. The FLAIR contextual string embeddings perform better than the word2Vec and network embedding models. We can also observe that AmFlair and MultFlairFT , which are trained and fine-tuned on our dataset, presents better results than the pre-trained MultFlair embeddings model. The XLM transformer-embedding could not produce meaningful predictions (all words are predicted as "Other"). The low performance reported indicates that NER for Amharic is a difficult task. This is due to the fact that named entities do not have distinctive characteristics such as capitalization. Named entities in Amharic are also derived mostly from proper nouns (**አበባ** - flower), from verbs (**አበራ** - shined), and from adjectives (**ጉብዜ** - clever).
- **Sentiment analysis**: For the sentiment analysis task, we have observed that the deep learning approach outperforms the different classical supervised classifiers. Unlike the NER and the POS tagging tasks, the deep learning approach based on the network embeddings, specifically the **Role2Vec** approach outperforms the other models. Based on our error analysis, we found out that sentiment analysis is challenging both for users and machines as the meaning of the tweet depends on a specific context. Moreover, metaphorical speech and sarcasm are very common in Amharic text, especially on the Twitter dataset, which makes automatic classification very difficult.

In general, we can see that the different semantic models impact various tasks. One semantic model will not fit the need of multiple NLP applications. Another observation is that fine-tuning models or building models with a corpus that is carefully crafted have a better impact on the specific tasks. We believe that the models we publish will help in the development of different NLP applications. It will also open a different research direction to conduct more advanced research as well as to carry out insightful analysis in the usage of semantic models for Amharic NLP.

## 5. Conclusions

In this work, we presented the first comprehensive study of semantic models for Amharic. We first surveyed the limited number of pre-trained semantic models available, which are provided as part of multilingual experiments. We built different semantic models using text corpora collected from various sources such as online news articles, web corpus, and social media texts. The semantic models we built include (1) word2Vec embeddings, (2) distributional thesaurus models, (3) contextualized string embeddings, (4) distributional thesaurus embedding obtained via network embedding algorithms, and (5) contextualized word embeddings. Furthermore, the publicly available pre-trained semantic models are fine-tuned using our text corpora.

We also experimented with five different NLP tasks to see the effectiveness and limitations of the various semantic models. Our experimental result showed that deep learning models trained with the different semantic representations outperformed the classical machine learning approaches. We publicly released all the nine semantic models, the machine learning models for the different tasks, and benchmark datasets to further advance the research in Amharic NLP .

**Author Contributions:** Conceptualization, S.M.Y.; methodology, S.M.Y. and A.A.A.; software, S.M.Y. and G.V.; validation, S.M.Y., A.A.A., I.G. and G.V.; formal analysis, S.M.Y., A.A.A. and G.V.; investigation, S.M.Y., I.G. and G.V.; resources, S.M.Y. and G.V.; data curation, S.M.Y., I.G. and G.V.; writing—original draft preparation, S.M.Y.; writing—C.B., S.M.Y. and A.A.A.; visualization, G.V.; supervision, C.B.; project administration, S.M.Y.; funding acquisition, C.B. All authors have read and agreed to the published version of the manuscript.

**Funding:** This research received no external funding.

**Data Availability Statement:** The resources such as benchmark NLP datasets for Amharic (PoS tagged dataset, NER annotated dataset, Sentiment dataset, Semantic similarity datasets), Preprocessing and segmentation tools, source codes for the model training, the Amharic corpus will be released in our GitHub repository (https://github.com/uhh-lt/amharicmodels (accessed on 24 October 2021)). The RoBERTa model will be published to the Hugginface repository (https://huggingface.co/uhhlt (accessed on 24 October 2021)). The FLIAR models will be released to the FLAIR list of public libraries.

**Conflicts of Interest:** The authors declare no conflict of interest.